\documentclass[twocolumn,american,5p, authoryear]{article}
\usepackage[T1]{fontenc}
\usepackage[latin9]{inputenc}
\usepackage{verbatim}
\usepackage{url}
\usepackage{graphicx}
\usepackage[authoryear]{natbib}

\makeatletter
\newcommand{\lyxaddress}[1]{
\par {\raggedright #1
\vspace{1.4em}
\noindent\par}
}

\makeatother

\usepackage{babel}
\begin{document}

\title{Evolving neural networks to follow trajectories of arbitrary complexity}

\author{Benjamin Inden$^{1}$ \& Jürgen Jost$^{2,3}$}
\date{\vspace{-5ex}}
\maketitle

\lyxaddress{$^{1}$Department of Computing and Technology, Nottingham Trent University,
United Kingdom}

\lyxaddress{$^{2}$Max Planck Institute for Mathematics in the Sciences, Leipzig
Germany}

\lyxaddress{$^{3}$Santa Fe Institute, Santa Fe, New Mexico, USA}
\begin{abstract}
Many experiments have been performed that use evolutionary algorithms
for learning the topology and connection weights of a neural network
that controls a robot or virtual agent. These experiments are not
only performed to better understand basic biological principles, but
also with the hope that with further progress of the methods, they
will become competitive for automatically creating robot behaviors
of interest. However, current methods are limited with respect to
the (Kolmogorov) complexity of evolved behavior. Using the evolution
of robot trajectories as an example, we show that by adding four features,
namely (1) freezing of previously evolved structure, (2) temporal
scaffolding, (3) a homogeneous transfer function for output nodes,
and (4) mutations that create new pathways to outputs, to standard
methods for the evolution of neural networks, we can achieve an approximately
linear growth of the complexity of behavior over thousands of generations.
Overall, evolved complexity is up to two orders of magnitude over
that achieved by standard methods in the experiments reported here,
with the major limiting factor for further growth being the available
run time. Thus, the set of methods proposed here promises to be a
useful addition to various current neuroevolution methods.

\textbf{This is the authors' version of the accepted manuscript, the
final and definite version can be found at \url{https://doi.org/10.1016/j.neunet.2019.04.013}.
This version is available subject to a Creative Commons 4.0 CC-BY-NC-ND
license.}
\end{abstract}

\section{Introduction}

Evolutionary algorithms have been used successfully to solve various
optimization problems including for scheduling, symbolic regression
in astronomy, optimizing antenna designs and shapes of car parts,
finding electronic circuits that perform a given function, and game
playing \citep{Poli2008,Eiben2015}. Similarly, neural networks have
many important applications, including recognition of speech and handwritten
digits, and robot control \citep{Schmidhuber2015}. Topologies and
connection weights of neural networks can be optimized by evolutionary
algorithms \citep{Floreano2008}, and the resulting networks can be
used to control robots, as has been done for some decades in the field
of evolutionary robotics \citep{Nolfi2000}. Nevertheless, there are
still very few real world applications of evolutionary robotics, and
progress towards more complex behaviors of interest seems to be slow
\citep{Doncieux2015,Pugh2016}.

For many of the earliest evolutionary robotics experiments, neural
networks with a fixed topology, i.e., a fixed number of nodes and
connections, were used \citep{Nolfi2000}. That way, the achievable
complexity of behavior is obviously limited. Later, methods were introduced
that could increase these numbers. Perhaps the most well known of
these is the method called NEAT \citep{Stanley2002}. This neuroevolution
method starts evolution using networks without any hidden nodes and
subsequently adds neurons and connections by carefully designed mutation
operators. It has been shown that complexification during evolution
does indeed occur when using NEAT, and can create neural networks
with in the order of up to a few dozens of neurons \citep{Stanley2004a}.
NEAT has subsequently been widely used for various evolutionary robotics
experiments. A more recent achievement is the development of methods
that can produce large neural networks from comparatively small genomes.
Some methods such as HyperNEAT \citep{Stanley2009} and Compressed
Network Complexity Search \citep{Koutnik2010,Gomez2012} achieve this
by evolving a process that constructs the neural network instead of
evolving the neural network directly. Other methods still evolve parts
of the network directly, but make use of user-specified constraints
(e.g., ICONE, \citet{rempis2012evolving}) or design patterns with
evolvable parameters (e.g., NEATfields, \citet{Inden2009a}) to create
larger networks from those parts. While evolution can produce comparatively
complex networks with hundreds of neurons using these methods, the
complexity of the behavior that can be produced by evolution (as opposed
to behavior that emerges from the coupling to a complex environment)
is still limited, with the increase of both complexity and fitness
typically converging towards zero after several hundred or thousand
generations.

We have previously argued \citep{Inden2013} that this is unavoidable
with the way artificial evolution is typically set up: A genome of
fixed length can only produce a limited number of behaviors (and therefore
there is an upper bound to complexity). A growing genome, on the other
hand, cannot grow indefinitely because either mutations will overpower
selection (in the case of fixed per-gene mutation rates, as discovered
by \citet{Eigen1971}) or the waiting times for mutations on individual
genes grow without bound (in the case of fixed per-genome mutation
rates). However, evolution can in principle get around this problem
by changing the genetic architecture such that mutations will with
a higher than uniform probability occur where they are needed. It
has actually been found that different regions in animal genomes are
subject to different mutation rates, and that this is under genetic
control \citep{Martincorena2012}. Evolution would be optimally adaptive
if previously evolved useful features of an organism were conserved
by a reduction of the local mutation rates, whereas features under
adaptive evolution had increased local mutation rates. The starting
point for this paper is a previously proposed method to guide mutations
towards features under active evolution, and away from previously
evolved adaptive features \citep{Inden2013}. The basic idea is that
only those parts of a neural network that were created by mutations
most recently can be mutated. Older structures are frozen and cannot
be mutated any more.

The idea of freezing evolved structure is not new, but has been explored
a number of times, typically in connection with modular neural networks
\citep{Huelse2004,Togelius2004}. Incrementality has also been a part
of traditional neural network architectures trained by supervised
learning such as Cascade-Correlation networks \citep{Fahlman1990}.
It has also played a key role within the evolving spiking neural network
models that were developed within a framework termed evolving connectionist
systems. In those models, the learning procedure typically adds new
neurons for new items of data to be learned without affecting existing
neurons much. The neural networks considered in that framework are
feed-forward networks (in some cases with additional structures such
as evolving feature selectors, or dynamic reservoirs) used for classification
tasks \citep{Schliebs2013}. Recently, approaches that use evolution
to learn a suitable architecture for a deep neural network (typically
a convolutional neural network) that is trained by supervised methods
have also become quite popular \citep{Real2017,MIIKKULAINEN2019293,Liu2018}.
This research has shown that evolved architectures can solve difficult
pattern recognition tasks with many inputs, classes, and samples,
and are often superior in that regard to hand-designed architectures.
However, as the architecture grows in size during evolution, two effects
can be expected: For one, the same convergence as described above
because the genome size increases. In addition, variance will increase
for the supervised learning tasks, which in turn increases the time
required for training the weights \citep{Geman1992}. Convolutional
neural networks use techniques such as weight sharing to mitigate
the latter problem. Recently, freezing of structure has also been
explored within this context. \citet{DBLP:journals/corr/RusuRDSKKPH16}
train columns of a deep neural network on different tasks one by one.
The connection weights in earlier evolved columns are frozen, but
lateral connections to newer columns are trained to allow for transfer
of skills between different tasks. Similarly, \citet{terekhov2015knowledge}
train a series of blocks of neural networks on a series of related
tasks, where the weights in earlier blocks are frozen. Both approaches
use neural networks that are feed-forward and have a fixed topology,
whereas evolutionary approaches can go beyond that.

If we want a sustained linear growth of complexity during evolution,
we must ensure that the properties of the environment in which evolution
takes place, including the population structure and the properties
of the neural networks that are relevant to the applied mutation operators,
remain constant on average over evolutionary time. We might call this
the stationarity property. Freezing old network structures is one
method that helps in achieving it. But as the investigations presented
here will show, it is not sufficient on its own and not even the major
contributor towards achieving that goal.

We study a task here where in its most basic version an agent must
follow a predefined trajectory on an infinite 2D plane for as long
as possible. This is not a difficult task as such: For holonomic robots,
the motion in each dimension can be directly calculated given a goal
point since it is independent from the motion in other dimensions.
For nonholonomic robots, there are also traditional methods to solve
the problem, or to calculate an approximate solution \citep{Laumond1998}.
Of course, the controller itself, a neural network, causes nonlinearities,
and might also cause dependencies between the signals produced for
different outputs. Newer research on trajectory learning features
prominently in the literature on robot programming by demonstration,
where one challenge is to calculate joint angles for a complex actuator
based on an observed trajectory. Hidden Markov models and other related
techniques are often used to tackle that challenge \citep{Vakanski2012,Field2016}.

In our version of trajectory following, the agent does not get any
information through its sensors on where it should be, but it dies
if it is too far from the trajectory, so evolution can adapt the agents
to follow that trajectory using open loop control. An early experiment
on robots without sensors was reported in \citet[section 5.4]{Nolfi2000},
although there was only a discrete choice between a fixed number of
spatial regions there. The task studied in \citet{Inden2013} required
navigation on a plane, but there were only two goal areas to navigate
between. General trajectory following is much more difficult for evolution.
We found (as discussed below) that the method introduced in \citet{Inden2013},
which was basically a NEAT algorithm with two new features, was not
sufficient to solve the task. In section \ref{sec:Methods}, we describe
an improved method that is. In section \ref{sec:Results-and-discussion},
we present results for trajectory following in 2D and 3D space, and
study the contribution of individual features of the method to its
overall performance. Finally, we discuss connections to the larger
context of evolutionary robotics research, and make suggestions on
how to use our method on a larger range of tasks, in section \ref{sec:Conclusions-and-further}.

\section{\label{sec:Methods}Methods}

\subsection{The trajectory following task}

A random trajectory starts at position $(0,0)$ and is generated as
follows: There are 16 possible directions with angles to the positive
x axis of between $\frac{\pi}{8}$ and $2\pi$. Every 30 time steps,
the trajectory randomly takes one of these directions. The distance
covered in a single step is 0.1. A number of random segments is generated
before evolution starts, but as individuals manage to follow the trajectory
for most of its length, new segments are added to it at the end. Besides
this elongation, the trajectory does not change over the course of
evolution.

The agent has no ordinary inputs besides a fixed bias input. It starts
at position $(0,0)$ and can move with a maximum speed of 0.2. It
has a holonomic drive (i.e., movement in both dimensions is controlled
independently). For each dimension, there are two outputs controlling
movement, if the first is above 0.0 and the second is not, then the
first output specifies the speed of movement in the positive direction,
whereas if the second output is above 0.0 and the first is not, then
the second output determines the movement in the negative direction.
If both outputs are positive, they block each other and no movement
occurs, the latter also happens if both outputs are negative. If the
agent is within a radius of 1.0 from the current trajectory point,
it accumulates a fitness of 1.0 minus its current distance. Otherwise,
its life is terminated. Therefore, the agent is required to follow
the trajectory closely over space and time.

\subsection{Three-dimensional extensions of the task}

The trajectory for the three-dimensional version is generated in the
same way as for the two-dimensional version except that there are
now 14 possible directions which correspond to the 8 corners of a
cube and the centers of its six surfaces. The agent starts at position
$(0,0,0)$ and can move with a maximum speed of 0.3. The holonomic
drive now needs six network outputs and works analogous to its 2D
counterpart. However, there is also a nonholonomic version of the
task where the first pair of outputs determines the movement speed
along the current orientation of the agent, whereas the second and
third pair of outputs control the orientation by rotation about the
z- and y-axis respectively. The angles of rotation can be changed
by at most $\pm\frac{\pi}{4}$ in a single step. The fitness function
is the same as in the two-dimensional version of the task.

\subsection{Neural networks}

The method presented here is partly based on the well known NEAT (NeuroEvolution
of Augmenting Topologies) method \citep{Stanley2002}, which has influenced
some design choices and parameters mentioned below. 

The activation $a_{i}(t)=\sum_{j\in J}w_{ij}o_{j}(t-1)$ of an individual
neuron $i$ at time $t$ is based on a weighted sum of the outputs
of the neurons $j\in J$ to which they are connected. The output of
a neuron is calculated by applying a sigmoid function on the activation:
$o_{i}(t)=\tanh(a_{i}(t))$. We also compare this with a variant where
the activation is calculated as $a_{i}(t)=\tau(\sum_{j\in J}w_{ij}o_{j}(t-1))+(1-\tau)a_{i}(t-1)$.
$\tau$ is a constant between 0 and 1 that is genetically specified
for each neuron individually. If $\tau=1$, the activation is wholly
determined by the current input and there is no difference to the
standard method, whereas for smaller $\tau$ values, the activation
is also dependent on past input and changes more slowly. This is a
simple implementation of a continuous time recurrent neural network.
These kinds of networks have been used successfully in a number of
evolutionary robotics experiments \citep{Beer1992,Beer2006}, and
one might think that they are better suited for the trajectory following
task because they can more easily generate temporal dynamics at different
time scales.

Connection weights are constrained to a range of $[-3,3]$ as is common
in previous work with NEAT. The rationale behind constraining weights
is that connection weights with a very large absolute value would
often lead to saturation of neurons with a sigmoid transfer function,
which in turn would limit the variety of output of the neural network.
The threshold value for all neurons is 0.0, but a constant bias input
is available in all networks.

All connections from neurons to network outputs have a weight of 1.
The value of that output is then calculated by applying a sine function
$\sin\pi x$ on the sum of all contributions. By virtue of its periodic
nature, the sine function is what we call a homogeneous output function
(one of the major contributions of this paper): No matter what the
current value of the output as determined by all current connections
from neurons is, a new connection can always change it to become any
value in its whole range $[-1,1]$. We compare this against applying
the $\tanh$ function on the input (this function is not a homogeneous
output function as input from existing connections could have brought
the value so far into one of its saturation areas already that a single
new connection could not change much), and against the approach used
in some NEAT implementations of just taking the mean value of all
contributions (this is also not a homogeneous output function, and
has the additional disadvantage that adding a new connection to an
output can change the output value even if the new connection does
not convey any signal (i.e., has a value of 0.0).

\subsection{Experimental setup and selection}

For each configuration, 20 runs with different random seeds are performed.
Each run lasts for 3500 generations and has a population size of 300.
Truncation selection is used for the majority of configurations, where
the best 5\% are selected for reproduction with mutation, although
one copy of each is kept unchanged (elitism).

In one experiment, we compare this simple method with a version of
the more elaborate speciation selection that NEAT uses by default.
Speciation selection protects innovation that may arise during evolution
against competition from fitter individuals that are already in the
population. As a prerequisite, globally unique reference numbers are
assigned to each connection gene once it arises by mutation, and are
used to calculate a distance measure between two neural networks.
The dissimilarity between two networks is calculated as $d=c_{r}r_{c}+c_{w}\sum\Delta w$,
where $r_{c}$ is the number of connections present in just one of
these networks, $\Delta w$ are the connection weight differences
(summed over pairs of connections that are present in both networks),
and the $c$ variables are weighting constants with $c_{r}=1.0$,
$c_{w}=1.0$ by default.

Using this dissimilarity measure, the population is partitioned into
species by working through the list of individuals. An individual
is compared to representative individuals of all species until the
dissimilarity between it and a representative is below a certain threshold.
It is then assigned to this species. If no compatible species is found,
a new species is created and the individual becomes its representative.

The number of offspring assigned to a species is proportional to its
mean fitness. This rather weak selection pressure prevents a slightly
superior species from taking over the whole population, and enables
innovative yet currently inferior solutions to survive. In contrast,
the selection pressure between members of the same species is much
stronger: the worst 60\% of the individuals belonging to that species
are deleted, after which the other individuals are selected randomly.
Species that have at least five individuals in the next generation
also take the best individual into the next generation without mutations.
If the maximum fitness of a species has not increased for more than
200 generations and it is not the species containing the best network,
its mean fitness is multiplied by 0.01, which usually results in its
extinction. Also, in order to keep the number of species in a specified
range, the dissimilarity threshold is adjusted in every generation
if necessary. This threshold adjustment mechanism is a slight difference
from the original NEAT speciation technique, and was introduced by
\citet{Green2006}. It seems to work well together with full elitism,
which the original speciation method did not use, but otherwise yields
very similar results. Here, the initial speciation threshold is 4.0,
and the target number of species is between 10 and 20. The numerical
parameters for speciation selection have been set based on previous
experiments with other tasks \citep{Inden2009a}. 

\subsection{Genetic representation and operators}

NEAT is a method for simultaneously evolving the topology and the
connection weights of neural networks. It starts evolution with one
of the simplest possible network topologies and proceeds by complexification
of that topology. More specifically, the common ancestor of the whole
population has one neuron for each output, each of which is connected
to all inputs. There are no hidden neurons. Here, we start with even
smaller networks: Each output has one neuron, and each of this neurons
is initially connected to 50\% of the inputs on average. It has been
shown previously that having evolution select inputs for the neural
network can result in superior performance as compared to starting
with a fully connected network if the input space is large \citep{Whiteson2005}.
During the course of evolution, further neurons and connections can
be added.

Our neuroevolution method mostly uses mutation operators that are
very similar to those of the original NEAT implementation for evolving
the contents of the field elements. Numerical parameters are set to
values used in previous experiments \citep{Stanley2002,Inden2009a,Inden2013}.
The most common operation is to choose a fraction of connection weights
and either perturb them using a normal distribution with standard
deviation 0.18, or (with a probability of 0.15) set them to a new
value. The application probability of this weight changing operator
is set to 1.0 minus the probabilities of all structural mutation operators,
which amounts to 0.929 in most experiments. A structural mutation
operator to connect previously unconnected neurons is used with probability
0.02, another to connect a neuron to an input or output is also used
with probability 0.02, while an operator to insert neurons is used
with probability 0.001. The latter inserts a new neuron between two
connected neurons. The weight of the incoming connection to the new
neuron is set to 1.0, while the weight of the outgoing connection
keeps the original value. The idea behind this approach is to change
the properties of the former connection as little as possible to minimize
disruption of existing functional structures. The former connection
is deactivated but retained in the genome where it might be reactivated
by further mutations. There are two operators that can achieve this:
one toggles the active flag of a connection and the other sets the
flag to 1. They are used with probability 0.01 each.

With a probability of 0.01, another mutation operator is used, the
addition of which to the set of operators constitutes another main
contribution of this paper. This operator creates a new neuron, and
connects it to one of the outputs. This neuron does not have any inputs
initially, therefore it does not change the network output. However,
it provides a starting point for evolution to create a new module.
Of course, there is the possibility that this neuron does not attract
any meaningful connectivity, ending up as redundant structure (bloat).
But as the experiments will show, that is a reasonable price to pay
given the vast increase in fitness.

For those experiments that use continuous time recurrent neural networks,
another operator, applied with a probability of 0.05, perturbs the
neurons' $\tau$ values using a normal distribution with standard
deviation 0.08. 

Once a new gene arises by mutation, it receives a globally unique
reference number. This number is generated by a global counter that
is incremented every time a new gene arises. The innovation numbers
are originally used by NEAT to align two genomes during the process
of recombination (although in the experiments reported here, no recombination
is used). Here, they are important for the method described in the
next section, and also for speciation selection in the experiment
that uses that selection method.

\subsection{Freezing of previously evolved structure}

Usually, mutations are applied on all genes with uniform probability.
In contrast, the method introduced in \citet{Inden2013}, which constitutes
another major contribution of our approach, allows mutations only
on the $c_{m}$ newest genes. Here we refine the previously used method
by considering neuron genes and connection genes separately, and allowing
mutations on the $c_{m}$ newest genes of either group. (This is done
to prevent problems in networks that have either very many new connections,
or unusually many new neurons. In such networks, it would not be possible
any more to mutate neurons, or connections, if their ages were not
considered separately.) The relative age of all genes is known because
their innovation numbers are ordered by the time of their creation,
so the smallest innovation numbers where mutations are still allowed
can be calculated at the beginning of the mutation procedure from
lists of all neuron / connection genes in the genome. $c_{m}$ should
obviously be greater than the number of neurons or connections in
the common ancestors. Based on this criterion, $c_{m}$ is set to
25 here. We found in preliminary experiments that the method is robust
to some variation in this parameter.

All mutations are forbidden on older genes, including perturbations
of the connection weights and changes of $\tau$ values. However,
connections from neurons that are positioned between $2c_{m}$ and
$c_{m}$ on the age sorted list to new neurons are allowed. This makes
it possible to connect newly evolved structures with older structures.
In our previous work \citep{Inden2013}, we allowed connections from
all old neurons regardless of age to new neurons, but we have changed
that as with growing number of neurons there would be an ever growing
number of neurons from which to choose, which could introduce the
very issues of non-stationarity that we want to avoid.

\subsection{Temporal scaffolding}

For the task considered here, the agent needs to move in different
directions at different times, therefore it needs to possess some
information that is correlated to the elapsed time. Given that neural
networks can generate internal dynamics, they could be expected to
track time entirely internally. However, this did not work very well
in our previous experiments \citep{Inden2013}. The method introduced
back then, in a modified form, constitutes another major component
of our approach. We provide a scaffolding input for each period of
time that has a value of 1.0 during its respective period, and 0.0
at other times. Other methods of providing this temporal information,
like providing the current round as a binary number on several inputs,
or providing some periodic inputs with different periods, were far
less effective in preliminary experiments \citep{Inden2013}. Providing
time as a single analog input would also not be expected to work well
as it would lead to saturation in neurons with sigmoid transfer function,
and because of the restricted range of connection weights. In the
experiments reported here, a period of activation for a given scaffolding
input corresponds to 20 time steps. The overall number of scaffolding
inputs is potentially unlimited, and it is growing linearly with the
time that organisms are able to survive. Obviously, these inputs need
to be presented to the evolutionary process in a sequential order
instead of all at once.

The mutation operators are the key components of the algorithm that
need to be provided with information on the input and output geometry
of the task to be evolved. In our implementation, this information
is just a list of network inputs and a list of network outputs. Each
input and output has a unique identification number. This number is
stored as a reference in the newly created connection gene whenever
a connection from input or to output is established. The neural networks
of the common ancestors are provided with 5 scaffolding inputs (active
during the first five periods of life time, and corresponding to input
IDs 1000 to 1004). Whenever a network lives longer than the time covered
by the currently available scaffolding inputs, 5 more scaffolding
inputs for later periods (with consecutive input IDs 1005, 1006, etc.)
are made available to the respective mutation operators. Whenever
a mutation is to be performed, the mutation operators check for the
newest scaffolding input to which the given neural network is actually
connected (i.e., the one with the highest ID $iid_{sup}$), and will
only create new connections from scaffolding inputs whose IDs are
larger than or equal to $iid_{sup}-5$ \textemdash{} in other words,
inputs no more than 5 positions before the highest current one.

Next to scaffolding inputs, the neural networks can also have a limited
number of normal inputs. In our implementation, all inputs whose ID
is below 1000 are treated as a normal input. Connections from these
can be established at any time. For the tasks presented here, the
only normal input is a bias input that is set to a fixed value of
1.0.

\section{\label{sec:Results-and-discussion}Results and discussion}

\subsection{Performance of the proposed method on trajectory following}

\begin{figure}
\includegraphics[scale=0.45]{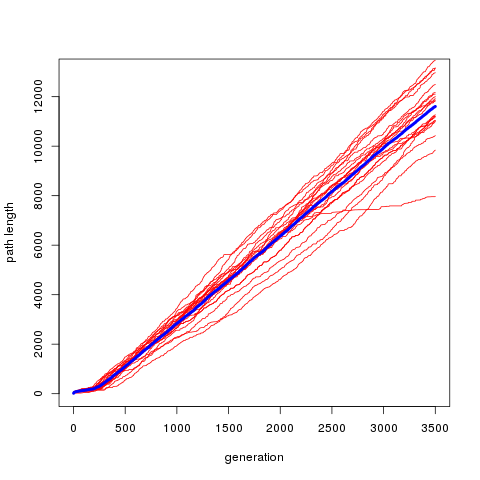}

\caption{\label{fig:2D}Performance of the standard configuration on 2D trajectory
following.}
\end{figure}
\begin{figure}
\includegraphics[scale=0.45]{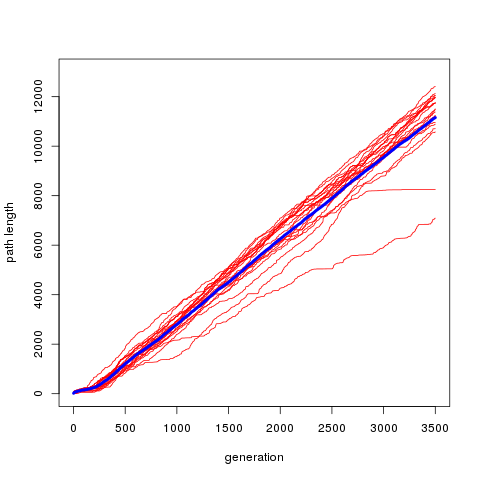}

\caption{\label{fig:2Dunequal}Performance of the standard configuration on
2D trajectory following with unequal lengths of segments.}
\end{figure}
\begin{figure}
\includegraphics[scale=0.45]{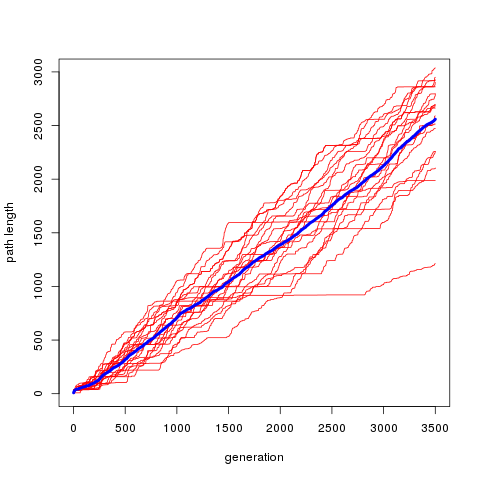}

\caption{\label{fig:3D}Performance of the standard configuration on 3D trajectory
following.}
\end{figure}
\begin{figure}
\includegraphics[scale=0.5]{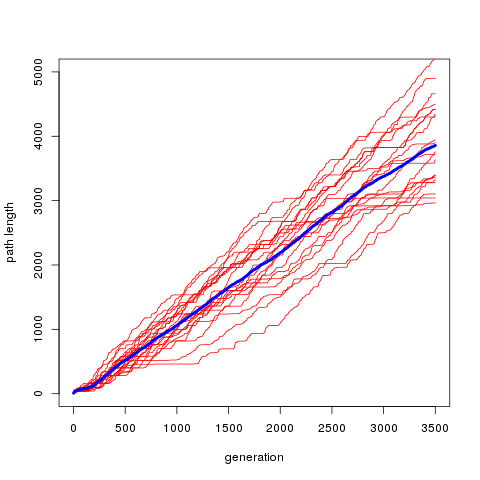}

\caption{\label{fig:3D-1}Performance of the standard configuration (population
size 1000) on 3D trajectory following.}
\end{figure}
\begin{figure}
\includegraphics[scale=0.45]{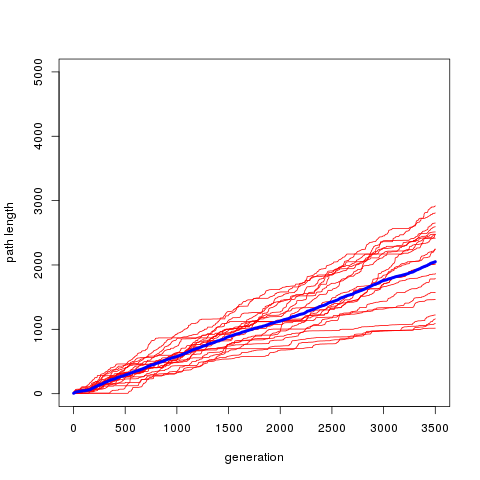}

\caption{\label{fig:3D-NH}Performance of the standard configuration on nonholonomic
3D trajectory following.}
\end{figure}

When the full set of methods described in the previous section is
applied to the 2D trajectory following task, a linear growth of the
length of the learned trajectory over evolutionary time can typically
be observed over the full 3500 generations. From Fig. \ref{fig:2D},
it is apparent that the mean growth of trajectory length is approximately
linear although in one run evolution switches to a lower speed at
some point. The fitness increase achieved within 50 generations is
$168\pm13$ at generation 1000 (where the second number is the standard
error, or uncertainty of the mean), $173\pm13$ at generation 2000,
and $188\pm16$ at generation 3000. There are no significant pairwise
differences between these values when using a Wilcoxon rank sum test,
so no significant deviation from linear growth can be found. Overall,
a trajectory length of $11604\pm286$ time steps is reached on average,
which corresponds to having adapted to about 387 segments, or 386
potential direction changes.

The fact that all path segments have a length of 30 time steps might
seem to simplify the task somewhat. If segment lengths are drawn from
a uniform distribution between 10 and 50, average trajectory length
reaches $11154\pm292$ (Fig. \ref{fig:2Dunequal}). This is not significantly
different ($p=0.31$, Wilcoxon rank sum test).

For the 3D trajectory following task, a performance of $2259\pm97$
is reached, which corresponds to 75 segments on average. It can be
seen from Fig. \ref{fig:3D} that the mean growth of trajectory length
is still approximately linear, but individual runs undergo periods
of stagnation, from which they recover sooner or later. By increasing
the population size to 1000 (Fig. \ref{fig:3D-1}), periods of stagnation
become shorter and the performance increases to $3858\pm147$. When
tackling the nonholonomic 3D trajectory following task using a population
size of 1000, a performance of $2051\pm137$ is reached (Fig. \ref{fig:3D-NH}).

These experiments make it plausible to assume that using the methods
presented here, it is indeed possible to evolve trajectories of arbitrary
length \textemdash{} or at least very long length \textemdash{} in
linear generation time. The observed periods of stagnation do not
seem to take away from the linearity as their frequency and duration
do not change in an obvious way over the observed number of generations.
Furthermore, it is possible to decrease stagnation by increasing the
population size. Further experiments (results not shown) using the
2D task and smaller population sizes confirm this as the frequency
and duration of stagnation periods increases in a similar way to the
way they change between Figs. \ref{fig:3D-1} and \ref{fig:3D}.

Because trajectory length grows approximately linearly, it makes sense
to calculate the speed at which evolution injects information about
the trajectory into the population as done previously \citep{Inden2015}.
As mentioned above, there are on average 386 decisions about the direction
learned within 3500 generations, and for each of them, there are 16
possible directions. So we have $\log_{2}16\cdot\frac{386}{3500}\approx0.44$
bit / generation. According to \citet{Kimura1961} and \citet{Worden1995},
the maximally attainable speed for a selection of 5\% should be $\log_{2}\frac{1}{0.05}\approx4.3$
bit / generation. Furthermore, that theory predicts that speed grows
logarithmically with selection strength, but is not correlated to
population size. We found in further experiments that a trajectory
length of $9340\pm251$ is reached for a selection of 10\%, and $6432\pm183$
for a selection of 20\%. Taking the logarithms of the speed ratios
for the 5\% vs. 10\% and 10\% vs. 20\% configurations, we get 0.31
and 0.54, respectively, which means that the speed grows less than
logarithmically with selection strength. From the experiments with
different population sizes mentioned above, it follows that there
is also a correlation between population size and speed. These observations
all indicate that the speed of evolution is limited by the availability
of suitable mutations rather than by selection in these experiments.
Increasing the population size obviously increases the number of mutated
offspring that a selected individual will have and therefore has a
direct influence on the speed limiting factor.

\subsection{The contribution of individual features to performance}

\begin{figure}
\includegraphics[scale=0.35]{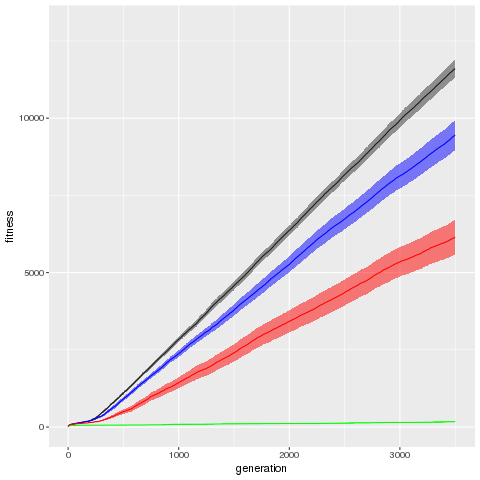}

\caption{\label{fig:Growth-of-trajectory1}Growth of trajectory length over
evolutionary time. Mean and uncertainty of the mean are shown for
a standard configuration (black), a configuration without freezing
of previously evolved structure (blue), a configuration without temporal
scaffolding (green), and a configuration without mutations that create
new connections to outputs (red).}
\end{figure}
\begin{figure}
\includegraphics[scale=0.35]{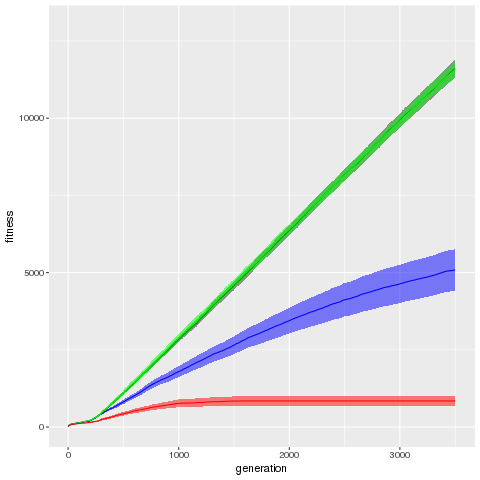}

\caption{\label{fig:Growth-of-trajectory2}Growth of trajectory length over
evolutionary time. Mean and uncertainty of the mean are shown for
a standard configuration (black, mostly behind green), a configuration
with tanh output function (blue), a configuration with mean output
function (red), and a standard configuration with the additional possibility
to have hidden nodes with sine transfer function (green).}
\end{figure}

\begin{figure}
\includegraphics[scale=0.35]{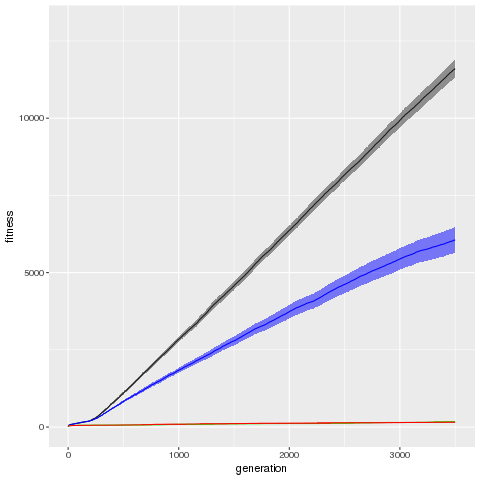}

\caption{\label{fig:Growth-of-trajectory-3New}Growth of trajectory length
over evolutionary time. Mean and uncertainty of the mean are shown
for a standard configuration (black), a CTRNN configuration (blue),
a standard configuration without temporal scaffolding (green, barely
visible behind red), and a CTRNN configuration without temporal scaffolding
(red).}
\end{figure}

\begin{figure}
\includegraphics[scale=0.35]{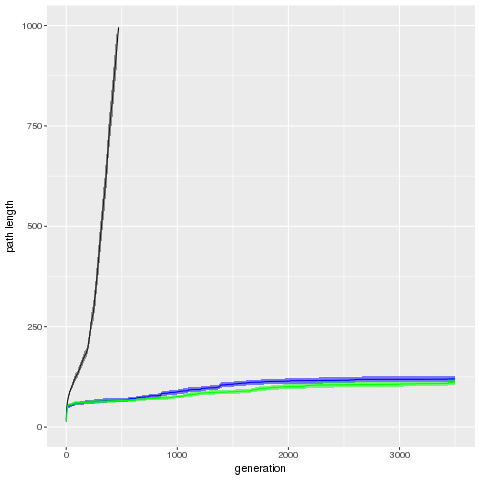}

\caption{\label{fig:Growth-of-trajectory4New}Growth of trajectory length over
evolutionary time. Mean and uncertainty of the mean are shown for
a standard configuration (black), a NEAT configuration with truncation
selection (blue), and a NEAT configuration with speciation selection
(green).}
\end{figure}

As argued in the introduction to this paper, freezing of previously
evolved structure should be an important factor enabling sustained
complexification. As shown in Fig. \ref{fig:Growth-of-trajectory1},
performance indeed becomes significantly worse without this technique
($p<0.001$, Wilcoxon rank sum test), but it is still at 81\% of the
original performance. Without mutations that create new pathways to
output, performance goes down to 53\%. Without temporal scaffolding,
there is barely any complexification at all.

Fig. \ref{fig:Growth-of-trajectory2} shows the importance of choosing
a homogeneous transfer function for the outputs. When using a hyperbolic
tangent function instead of a sine function, average performance goes
down to 44\%. If the mean function is used, average performance drops
to 7\%. If evolution can use the sine transfer function for hidden
nodes in addition to for outputs, there is no significant difference
to the standard method ($p=0.81$, Wilcoxon rank sum test). This supports
the idea that the advantage of using the sine function is really in
avoiding output saturation and not in any increased evolvability conferred
by using the function as such.

Fig. \ref{fig:Growth-of-trajectory-3New} shows results on methods
to generate the temporal dynamics necessary for trajectory following
in the neural networks. For this task and with the technical details
as specified in this paper, no advantages can be found for using continuous-time
recurrent neural network, i.e., networks where the individual neurons
change their states slowly as determined by evolved time constants.
When used with temporal scaffolding, adding this feature leads to
a drop in average performance to 52\%, while without temporal scaffolding,
both normal and continuous-time networks perform badly with no significant
difference ($p=0.63$).

Fig. \ref{fig:Growth-of-trajectory4New} shows how the method proposed
here compares to standard NEAT and a variant thereof that uses the
same strong truncation selection as the standard configuration. These
reach performances of $111\pm6$ and $120\pm7$, respectively. This
means that using all the methods presented here increases the performance
on the trajectory following task by a factor of 105 as compared to
standard NEAT within 3500 generations. Because NEAT leads to stagnation,
this factor can be assumed to increase if evolution lasts for more
generations.

\begin{figure}
\includegraphics[scale=0.35]{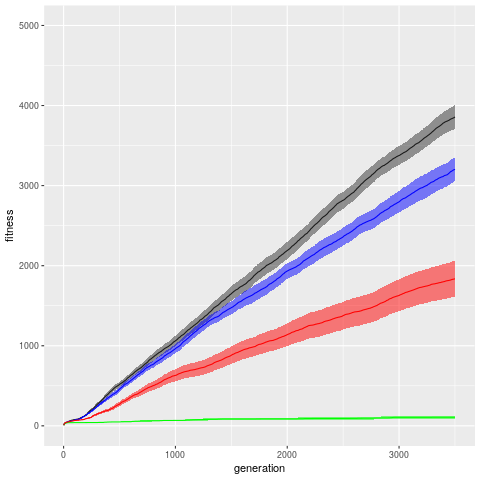}

\caption{\label{fig:Growth-of-trajectory1-1}Growth of trajectory length for
the 3D task over evolutionary time. Mean and uncertainty of the mean
are shown for a standard configuration (black), a configuration without
freezing of previously evolved structure (blue), a configuration without
temporal scaffolding (green), and a configuration without mutations
that create new connections to outputs (red).}
\end{figure}
\begin{figure}
\includegraphics[scale=0.35]{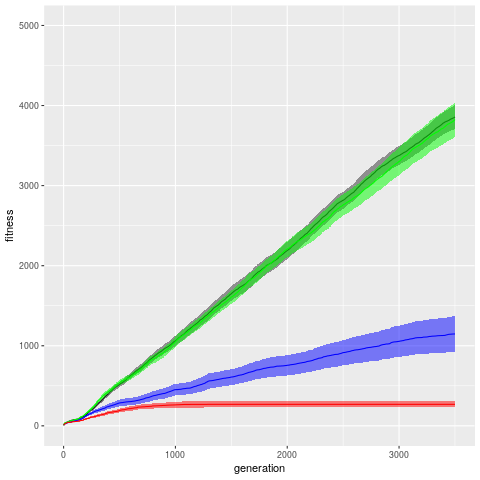}

\caption{\label{fig:Growth-of-trajectory2-1}Growth of trajectory length for
the 3D task over evolutionary time. Mean and uncertainty of the mean
are shown for a standard configuration (black, mostly behind green),
a configuration with tanh output function (blue), a configuration
with mean output function (red), and a standard configuration with
the additional possibility to have hidden nodes with sine transfer
function (green).}
\end{figure}

\begin{figure}
\includegraphics[scale=0.35]{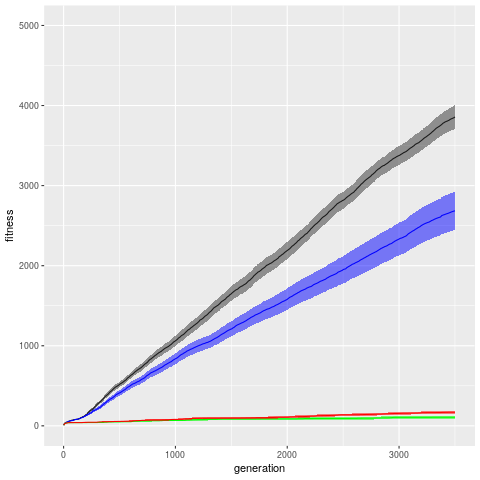}

\caption{\label{fig:Growth-of-trajectory-3New-1}Growth of trajectory length
for the 3D task over evolutionary time. Mean and uncertainty of the
mean are shown for a standard configuration (black), a CTRNN configuration
(blue), a standard configuration without temporal scaffolding (green),
and a CTRNN configuration without temporal scaffolding (red).}
\end{figure}

\begin{figure}
\includegraphics[scale=0.35]{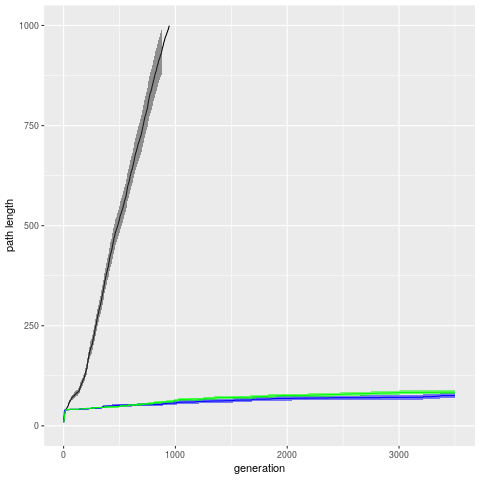}

\caption{\label{fig:Growth-of-trajectory4New-1}Growth of trajectory length
for the 3D task over evolutionary time. Mean and uncertainty of the
mean are shown for a standard configuration (black), a NEAT configuration
with truncation selection (blue), and a NEAT configuration with speciation
selection (green).}
\end{figure}
Finally, as shown in Figs. \ref{fig:Growth-of-trajectory1-1} \textendash{}
\ref{fig:Growth-of-trajectory4New-1}, similar conclusions regarding
the contribution of individual features of the method can be drawn
if the experiments are repeated for the 3D trajectory following task
and a population size of 1000.

\subsection{Structure and function of an evolved neural network}

\begin{figure*}
\begin{centering}
\includegraphics[scale=0.1]{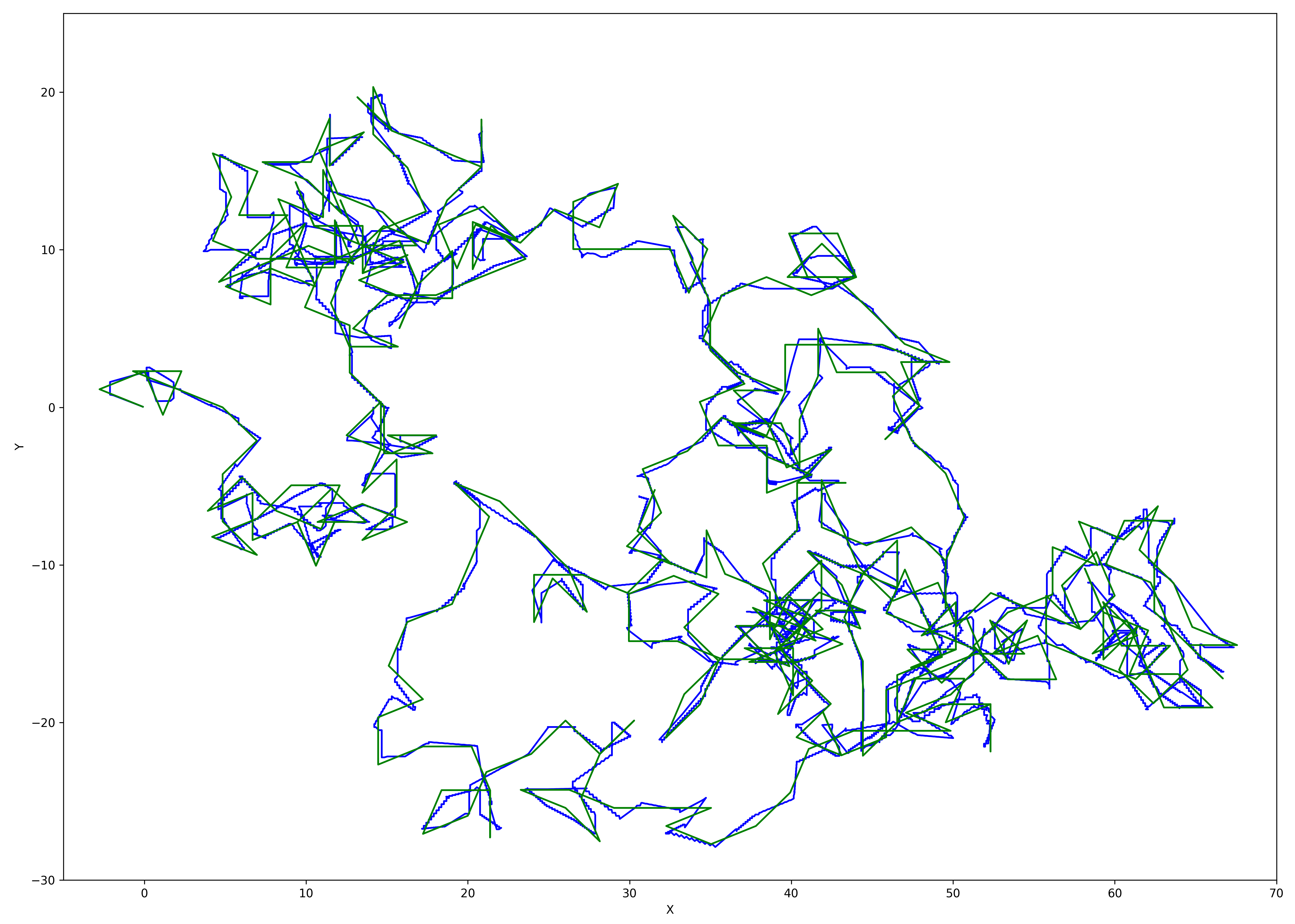}
\par\end{centering}
\caption{\label{fig:Trajectory-highest}Trajectory of the highest performing
individual from the best run. Green: optimal trajectory; blue: actual
trajectory. Both start at (0, 0).}
\end{figure*}
\begin{figure*}
\begin{centering}
\includegraphics[scale=0.32]{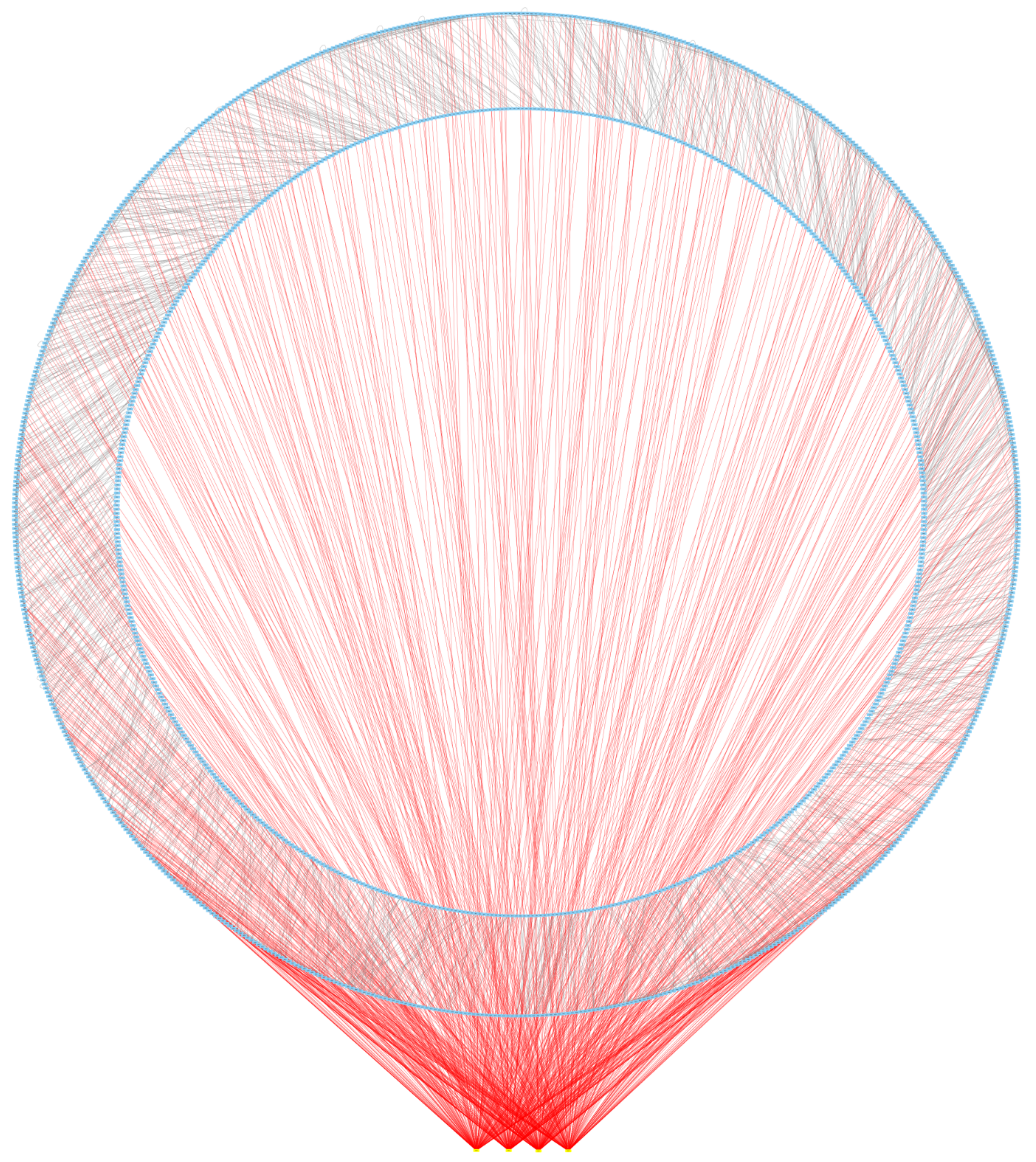}
\par\end{centering}
\caption{\label{fig:The-topology-highest}The topology of the highest performing
individual. Inner ring: inputs; outer ring: hidden neurons. Outputs
are shown at the bottom. Connections to outputs are shown in red,
while all other connections are drawn in gray. A spoke-like pattern
between inputs and hidden neurons and an absence of long range connections
are the dominant visible features at that scale. Image created using
Cytoscape \citep{shannon2003cytoscape}.}
\end{figure*}
\begin{figure*}
\begin{centering}
\includegraphics[clip,scale=0.17]{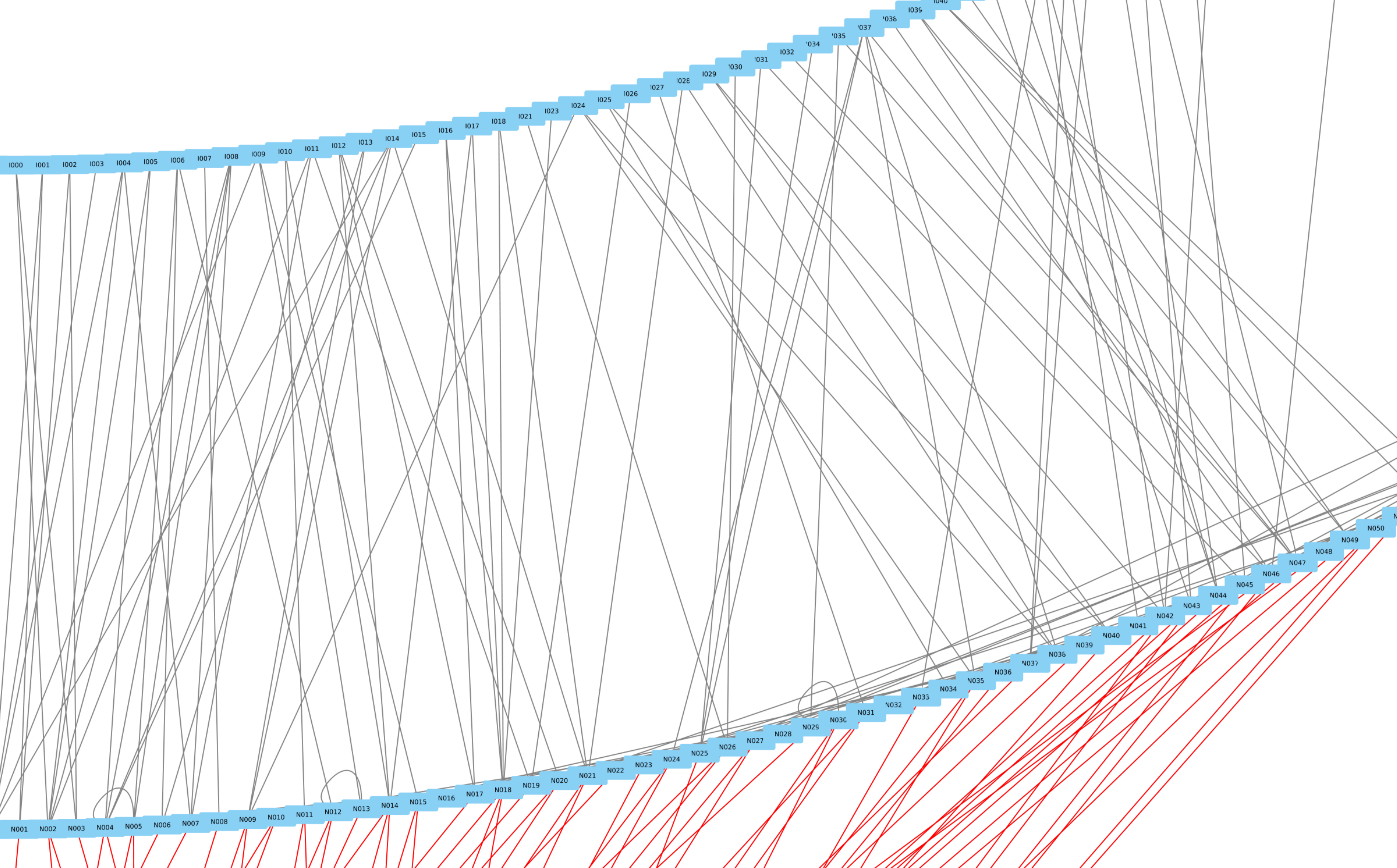}
\par\end{centering}
\caption{\label{fig:Magnification-of-topology}Magnification of a part in the
lower right section of the network from Fig. \ref{fig:The-topology-highest},
starting with the first hidden neuron and then proceeding anticlockwise.
Some recurrent connections as well as medium range connections between
neurons are visible at this scale.}
\end{figure*}
\begin{figure}
\includegraphics[scale=0.5]{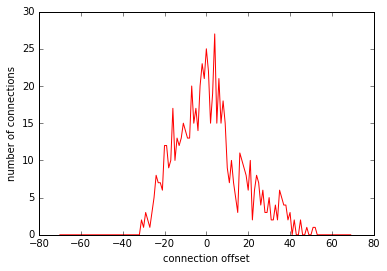}

\caption{\label{fig:Number-of-connections}Number of connections in the highest
performing individual that go from a neuron to another neuron, where
the neuron numbers are assigned according to historical order (i.e.,
when they arose by mutation).}
\end{figure}
\begin{figure}
\includegraphics[scale=0.5]{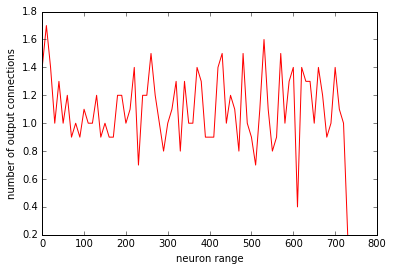}

\caption{\label{fig:Connections-output}The average number of connections going
from neurons (sorted in historical order and averages taken over groups
of 10 neurons each) to outputs for the highest performing individual.}
\end{figure}

The trajectory of an evolved individual from the best run can be seen
in Fig. \ref{fig:Trajectory-highest}. With the way the fitness function
is designed, the individual can get off the original trajectory as
long as it is not too far. If a higher accuracy than that is desired,
it is possible to scale the whole task, i.e., decrease both the survival
radius and the speed with which the individuals and the target point
can move. That way, higher accuracy would be achieved at the expense
of speed. If only the survival radius is decreased, the task will
get more difficult. We expect that this could be compensated for by
a larger population size.

The network of the best individual consists of 676 network inputs
(675 of which are scaffolding inputs), 733 neurons, 4 outputs, 1301
connections from input to neurons, 710 connections between neurons,
and 816 connections to output. Given that this particular neural network
has been adapted to follow 449 segments of the trajectory, it follows
that on average 1.6 neurons are used per segment. The low connection
to neuron ratio suggests that the network consists of relatively independent
modules. Indeed, if network inputs and outputs are removed, the remaining
network has 139 connected components, with 5.3 neurons per component
on average. This is due to the rareness of medium range connections
and absence of long range connections between neurons, which can be
easily seen in Figs. \ref{fig:The-topology-highest} and \ref{fig:Magnification-of-topology}.
Fig. \ref{fig:Number-of-connections} confirms the visual impression
by providing the distribution of internal connections within the network.
Also, given that a scaffolding input is active for 20 time steps and
a segment for 30 time steps, we would expect $\left\lceil 449\cdot\frac{30}{20}\right\rceil =674$
scaffolding inputs to be present, with a slightly higher number being
possible because 5 scaffolding inputs are added to the system in one
step whenever necessary as described above. The evolved network conforms
to this expectation.

Fig. \ref{fig:Connections-output} shows that the number of connections
to outputs does not change with the evolutionary age of a neuron except
for the very oldest and very youngest neurons. The former exception
might be related to the slower speed of evolution in the first few
hundred generations (Fig. \ref{fig:2D}), whereas the latter is because
those parts of the network are still under active evolution. In summary,
the network structure shows that our methods have achieved the desired
stationarity property to a high degree.

\section{\label{sec:Conclusions-and-further}Conclusions and future work}

One conclusion following from the experiments is that the types of
neural networks that are often used in evolutionary robotics experiments
are not very evolvable. Adding a new kind of mutation and changing
the transfer functions of outputs brings some improvement, but the
greatest improvement occurs when scaffolding is used. It seems that
standard neural networks are not good at producing varied behavior
over time \textemdash{} at least not in a way that is easily accessible
for evolution. That raises the question whether there might be any
sets of building blocks that are more evolvable than neural networks
and at the same time reasonable abstractions of biological building
blocks. Previous work has used spiking neural networks, compositional
pattern producing networks, and finite state automata among others
as alternative models \citep{DiPaolo2002,Floreano2005,florian2006spiking,Stanley2007,Riano2012}.
As was demonstrated here, using neural networks augmented by scaffolding
might be another option. One could argue that various external and
internal processes could provide scaffolds for evolution in natural
environments, which are typically much richer than artificial environments
\citep{Banzhaf2006}.

The evolutionary process achieved here is one of sustained rapid complexification
without any branching. This can be very useful for technical applications,
but we would not argue that this is the normal mode of evolution.
Indeed, various authors have argued that there is no general driven
trend towards more complexity during evolution \citep{McShe1996,Miconi2008}.
If anything, one might think that such a dynamics could arise locally
for a limited time. Obviously, total freezing of previously evolved
structure is also not a biologically plausible assumption. It might
be interesting to try to find possible biologically more plausible
mechanisms that could approximate the dynamics studied here for some
evolutionary time using a task similar to the one described here as
a benchmark.

As argued in \citet{Inden2012}, the evolutionary process set up here
can be argued to be an instance of open-ended evolution according
to the formal definition given by \citet{Bedau1998}. However, since
that early work, the scientific debate on open-ended evolution has
moved on. Many authors now consider the emergence of new evolutionary
units and levels, and/or the continuing presence of highly diverse
solution candidates, as hallmarks of open-ended evolution \citep{Taylor2016,Banzhaf2016}.
We have not addressed those issues here, which is why we would describe
the observed dynamics as sustained complexification instead.

The paths to be learned here are essentially random sequences, therefore
evolution proceeds towards behaviors with higher algorithmic or Kolmogorov
complexity \citep{Li1997}. Many researchers are more concerned with
whether the complexity concerned with structural regularities can
increase during evolution \citep{Ay2011}. We do not address this
question here. The growth of algorithmic complexity of the genotype
and phenotype seems to be a necessary, but not sufficient, condition
for the evolution of interesting complex behaviors. The point here
is not to argue that any task of interest might require that the complexity
of solutions grows forever. It is rather that many problems of interest
seem to require solutions of an algorithmic complexity that is unknown
in advance, but most likely higher than what can be achieved with
current methods. For these problems, techniques that can achieve a
sustained growth of complexity should be useful.

A related concern is that learning normally entails the ability to
solve previously unseen tasks that are similar to tasks seen during
training. Our evolved solutions do not have this ability in general.
For example, imagine a new path where only the direction of a single
segment would be changed. This could be sufficient to increase the
distance between the individual and its path so much that it would
die, rendering all skills for subsequent path segments useless. Furthermore,
because the structure bringing about the movement at this time would
likely already be frozen, evolution would not be able to re-adapt
the individual within a few generations. Therefore, it is probably
more accurate to say that evolution adapts individuals towards following
the path instead of saying that the path has been learned. That said,
it seems that much of the generalization ability, or ability to perform
well despite slight environmental change, in natural organisms results
either from developmental processes that are dependent on the environment
\citep{Sultan2007}, or from learning mechanisms within their neural
networks \citep{Downing2007a}. Therefore, we suggest that the proper
role of artificial evolution in creating robot controllers might be
to build structures that are capable of both performing some rather
fixed action patterns (of which the tasks studied here provide abstract
instances) and learning more advanced and flexible behavior based
on these skills during their lifetime. Unsurprisingly, the evolution
of plastic neural networks is an active research area \citep{Floreano2008,Soltoggio2008,Tonelli2013}.
The concept of a fixed action pattern was introduced in Lorenz \citeyearpar{Lorenz1935,Lorenz1981}.
It refers to a complex spatio-temporal pattern of behavior that arises
as an instinctive response to a stimulus. Once elicited, it typically
lasts much longer than the stimulus and executes independently of
further external input. Examples include various escape behaviors,
courtship and aggression signals, and ecdysis related behavior in
insects. Although the concept has been criticized in later research
with respect to the assumed full independence from external stimuli
in some of the mentioned examples, it continues to be a useful concept
in both biology and robotics \citep{Lawrence2002,Watt2003,Borst214,Kim2015,Prescott2006}.
Our methods can create fixed action patterns of high complexity. If
the scaffolding inputs are made dependent on the environment, these
action patterns can even become flexible to a degree.

From a technical perspective, one might be most interested in the
question of how well the method can be used on other tasks. The tasks
studied in this article represent tasks that require a complex fixed
pattern of movement to be executed. This would also include trajectories
in non-physical spaces, for example the generation of complex sound
patterns, or the execution of a developmental program by a simple
model of a genetic regulatory network in order to create a complex
phenotype \citep{Hinman2003,Stanley2003,Doursat2013}. One possible
criticism regarding limitations of the method would be to think of
a robotics-related version of a trap fitness function, where following
a certain path is required, but once a certain point has been reached,
the path to be followed earlier in order to achieve the highest fitness
changes. These kinds of tasks could not be solved by our method in
its pure form. However, one needs to consider three points: (1) It
is an open question whether any of these tasks would actually correspond
to any useful instinctive behavior that can be observed in natural
organisms, or would be useful for artificial organisms. Evolution
cannot solve all kinds of problems anyway \citep{Barton2000}, and
incrementality \textemdash{} building new function on top of old function
instead of changing the old function \textemdash{} is a key feature
of many evolved solutions. (2) Some kinds of these problems might
be successfully tackled by using diversity preserving selection mechanisms,
perhaps together with recombination, and developing these is an orthogonal
research direction in evolutionary robotics \citep{Pugh2016}. The
resulting methods could possibly be combined with the methods presented
here. (3) While we freeze all but the most recently evolved structure
here, an experimenter with some prior understanding of the task to
be solved could inject that knowledge into the evolutionary process
by freezing different parts of the structure at different evolutionary
times.

The tasks studied here, and the task studied in \citet{Inden2013},
are all tasks that require certain behavior at certain points in time.
Other tasks might require certain behaviors in certain situations,
where situations might occur repeatedly and be determined by various
factors in the environment. In that case, one could still use the
same methods, including the incremental provision of inputs, but just
use inputs that indicate the situation instead of the time. This input
could come from sensors or even from subnetworks that have previously
been evolved to recognize certain situations.

The experiments on trajectory following in higher dimensional spaces
seem to indicate that performance goes down as the number of outputs
increases. This is another limitation of the presented method. It
is understandable given the large increase in local search space that
the addition of only one output incurs (in principle, all neurons
could be connected to that output). If there are more than just a
few outputs, then perhaps one can use incremental provision of outputs
in a similar fashion to the way inputs are provided incrementally
if the task allows to learn their behavior separately. Of course,
tasks with many outputs have been previously solved using methods
that exploit the geometric regularities of the tasks \citep{Stanley2009,Inden2009a}
\textemdash{} another orthogonal research direction whose results
could be easily combined with the methods presented here. For example,
large and regular HyperNEAT networks are generated by an underlying
network \textemdash{} one could apply the methods discussed here onto
that underlying network. In NEATfields, applying our methods would
also be straightforward; the main task would be to specify how those
additional mutation operators that change the parameters of the design
patterns interact with the freezing of previously evolved structure.

Another limitation of the approach presented here is that while the
number of necessary generations depends linearly on the trajectory
length, the run time of the algorithm is at least quadratic. That
is because the maximum and average lengths of trajectories achieved
before individuals die also grow linearly over the generations. (An
even higher computational complexity is expected because the size
of the neural networks also increases over time.) With our implementation
and available hardware, run times of up to two weeks were typical
for standard configurations. That said, speedups could be achieved
by parallelizing fitness evaluation. It would also be possible to
divide a longer trajectory into several shorter ones, evolve trajectory
following for each one separately, and then evolve or hand design
a mechanism that ensures that the right subnetworks are active at
each time step. If the length of the individual pieces of the trajectory
remained constant and ignoring any effort for the arbitration mechanisms,
that would result in a linear run time. 

Based on these thoughts and the results of the experiments presented,
we conclude that the methods presented here \textemdash{} freezing
of previously evolved structure, temporal scaffolding, a homogeneous
transfer function for output nodes, and mutations that create new
pathways to outputs \textemdash{} might become essential ingredients
of future methods that are used to evolve neural networks for robot
control and other tasks.

\subparagraph{Declarations of interest:}

none. This research did not receive any specific grant from funding
agencies in the public, commercial, or not-for-profit sectors.

\bibliographystyle{apalike}
\bibliography{../benjamin}

\end{document}